\documentclass[journal]{IEEEtran}

\usepackage{amsmath,amssymb,amsthm}
\usepackage{graphicx}
\usepackage{booktabs}
\usepackage[hypertexnames=false]{hyperref}
\usepackage{microtype}
\usepackage{mathtools}
\usepackage{xcolor}
\usepackage{array}
\usepackage{enumitem}
\usepackage{algorithm}
\usepackage{algpseudocode}
\usepackage{tikz}
\usetikzlibrary{positioning}

\hypersetup{colorlinks=true, linkcolor=blue, citecolor=blue, urlcolor=blue}

\definecolor{darkblue}{RGB}{30,58,95}
\definecolor{medblue}{RGB}{46,95,163}
\definecolor{lightblue}{RGB}{58,123,200}
\newtheorem{definition}{Definition}

\title{A Hierarchical Consistency Framework for Auditing Retrieval-Augmented Generation Systems}
\author{Ram\'on Gonz\'alez and Antonio D\'iaz\\
Mentomy AI\\
\texttt{ramon@mentomy.com}}

\begin{document}
\maketitle
\pagestyle{plain}
\thispagestyle{plain}

\begin{abstract}
Retrieval-augmented generation (RAG) is commonly evaluated by whether
the final answer is correct. That test is insufficient: an answer can
match its reference while the context that produced it contains a
direct contradiction, leaving the contested evidence invisible to
answer-only review and retrieval-relevance scores. This paper presents
the Hierarchical Consistency Framework (HCF), a post-hoc,
model-agnostic audit of three distinct levels of a RAG process: the
knowledge corpus, the final retrieved context, and the generated
answer. HCF represents corpus conflicts as source-linked atomic facts,
thereby identifying the documents responsible, and returns each Answer
Consistency Score (ACS) with an explanation of supporting and
contradictory contextual statements. We evaluate HCF on several controlled
corpora spanning five domains and 100 query--corpus instances. A human
evaluator compares every generated response with its supplied
ground-truth response. The results show that the three diagnostic levels
can dissociate: the corpus with the highest mean retrieval similarity
has the lowest mean ACS, while a structurally degraded corpus performs
worse at corpus level but better at answer level. Most importantly,
HCF identifies contradictory retrieved evidence in several cases where
the answer still matches the ground truth. HCF does not certify factual
truth; it makes the evidence supporting and challenging an answer
inspectable and attributable.
\end{abstract}

\begin{IEEEkeywords}
Retrieval-augmented generation, consistency auditing, corpus quality,
conflicting evidence, retrieval diagnostics, answer grounding,
explainable artificial intelligence.
\end{IEEEkeywords}

\section{Introduction}\label{sec:introduction}\label{sec:intro}

Large language models (LLMs) encode vast world knowledge in their
parameters, yet this knowledge is static, opaque, and prone to
factual errors on facts outside the training distribution
\cite{lewis2020rag}. Retrieval-Augmented Generation (RAG) was
designed to address these limitations by augmenting a parametric
generator with a dynamic, non-parametric memory component: at
inference time, a retriever selects relevant evidence from an
external corpus and conditions generation on that evidence
\cite{lewis2020rag}. This architecture has become the dominant
production paradigm for knowledge-intensive tasks, including
enterprise search and question answering \cite{akkiraju2024facts}.
It extends a long line of retrieval-based question-answering systems,
from domain-specific symbolic systems to modern open-domain neural
architectures \cite{green1961baseball,woods1972lunar,zhu2021retrieving}.

Despite this success, diagnosing unreliable RAG outputs remains
challenging because the cause may originate at any of three levels:
the corpus may
contain structurally degraded or factually inconsistent documents;
the retriever may surface ambiguous or contradictory evidence; or
the generator may fail to ground its answer in the retrieved
context. Dense retrieval scores quantify query--passage semantic
relevance rather than factual agreement across sources. Consequently,
topically similar passages containing conflicting claims may receive
comparable scores, making retrieval similarity insufficient on its own
for diagnosing cross-document inconsistency
\cite{chen2022conflicts,wang2025conflicting}.

This paper formalizes consistency diagnostics through a
\emph{Hierarchical Consistency Framework} (HCF) for RAG systems,
organized across three levels: Corpus Consistency (CC), Retrieval
Consistency, and Answer Consistency. The three levels decompose the
audit and localize distinct failure modes; they are not assumed to have
a deterministic or monotonic relationship. The framework provides a measurable,
post-hoc, model-agnostic diagnostic at each level and is evaluated on a
controlled corpus in which intra- and inter-document degradation are
varied systematically. The retrieval-level metrics
($SS_1$, $SS_\mu$, and $SS_\sigma$) and the Answer Consistency Score
(ACS) also form an explainability layer that helps localize failures
across the RAG pipeline, directly contributing to the agenda of
Explainable AI (XAI) \cite{arrieta2020xai}.

\section{Related Work}\label{sec:related_work}\label{sec:related}

This section positions HCF relative to research on RAG evaluation,
corpus quality, conflicting evidence, and explainable AI.

RAG combines neural retrieval with generation over external
collections \cite{lewis2020rag}.
Dense passage retrieval, retrieval-augmented pre-training, and
fusion-in-decoder established the principal modern architectures
\cite{karpukhin2020dpr,guu2020realm,izacard2021leveraging}. Production
analyses identify corpus quality as a major RAG failure point
\cite{barnett2024seven}, motivating corpus-aware diagnostics across
enterprise, educational, and clinical deployments
\cite{neupane2024chatbot,hetz2024urobot}.

\textbf{Intra-document quality and data-centric AI.} Information-quality
research treats accuracy, completeness, consistency, and timeliness as
distinct assessment dimensions
\cite{redman1996data,naumann2000quality,pipino2002data}. Data-centric AI
similarly emphasizes systematic data improvement rather than model-only
optimization \cite{motamedi2021datacentric,zha2023datacentric}. We operationalize
these principles for RAG through automated diagnostics of document
quality and cross-document consistency.

\textbf{Inter-document conflict and conflicting evidence.} Entity-swap
studies show that factual conflicts can increase hallucination while
preserving fluency and topical relevance, making conflicting documents
difficult to distinguish through similarity-based retrieval
\cite{longpre2021entity,wang2025conflicting}. Under contradictory
evidence, models also vary in whether they follow retrieved context or
parametric knowledge \cite{chen2023rich,xie2023adaptive}. IndexRAG performs complementary
index-time analysis by extracting atomic knowledge units for
cross-document reasoning \cite{bao2026indexrag}. HCF instead uses
source-linked atomic fact triples to quantify cross-document conflict
before retrieval. This similarity-blindness motivates IeDCD as a
corpus-level diagnostic and $SS_\sigma$ as a query-level indicator of
retrieval discrimination.
Instruction-based context validation provides another closely related
line of work, but it generally evaluates the retrieved context or answer
without the source-linked corpus decomposition proposed here
\cite{gokul2025contradiction}.

\textbf{Explainable AI for RAG systems.} Post-hoc, model-agnostic
explanations operate without access to model internals
\cite{arrieta2020xai,ribeiro2016lime}, an important property
for RAG because retrieval and generation jointly determine the output.
HCF localizes failures through decomposable corpus diagnostics,
retrieval-score dispersion, and an answer score with a textual
justification. This connects $SS_\sigma$ to semantic-uncertainty
diagnostics \cite{kuhn2023semantic} and ACS to process-transparent XAI,
addressing calls for deployable explanation methods
\cite{arrieta2020xai,ali2023xai,longo2024xai2}.

\section{Methodology}\label{sec:problem_statement}\label{sec:method}

This section defines the Hierarchical Consistency Framework (HCF) as
an implementation-agnostic mathematical model of corpus, retrieval,
and answer consistency. The estimators used to instantiate these
quantities in the experiments are specified separately in
Section~\ref{sec:experimental_design}.

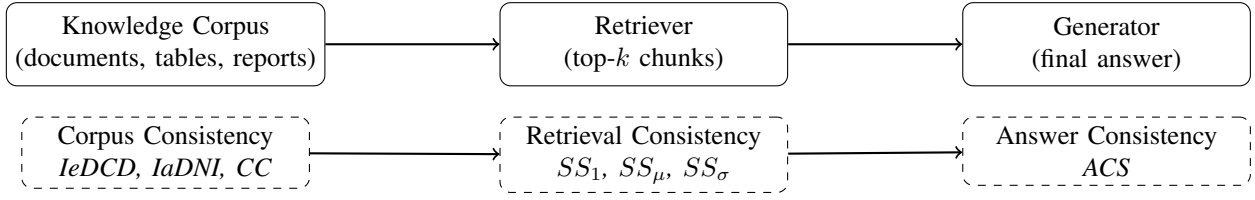
\begin{figure*}[!t]
\centering
\begin{tikzpicture}[
    node distance=1.6cm,
    box/.style={rectangle, draw, rounded corners, align=center, minimum width=3.8cm, minimum height=1.1cm},
    metric/.style={rectangle, draw, dashed, rounded corners, align=center, minimum width=3.8cm, minimum height=0.9cm},
    arrow/.style={->, thick}
]

\node[box] (corpus) {Knowledge Corpus\\(documents, tables, reports)};
\node[metric, below=0.4cm of corpus] (cc) {Corpus Consistency\\\textit{IeDCD, IaDNI, CC}};

\node[box, right=2.3cm of corpus] (retrieval) {Retriever\\(top-$k$ chunks)};
\node[metric, below=0.4cm of retrieval] (rc) {Retrieval Consistency\\\textit{$SS_1$, $SS_\mu$, $SS_\sigma$}};

\node[box, right=2.3cm of retrieval] (answer) {Generator\\(final answer)};
\node[metric, below=0.4cm of answer] (ac) {Answer Consistency\\\textit{ACS}};

\draw[arrow] (corpus) -- (retrieval);
\draw[arrow] (retrieval) -- (answer);

\draw[arrow] (cc) -- (rc);
\draw[arrow] (rc) -- (ac);

\node[align=center, above=0.3cm of retrieval] {\textbf{Diagnostic decomposition across the RAG process}};

\end{tikzpicture}
\caption{Overview of the proposed hierarchical consistency framework. Corpus Consistency characterizes the quality and agreement of the knowledge base; Retrieval Consistency characterizes the coherence of the retrieved evidence; and Answer Consistency characterizes the grounded agreement between the generated answer, the retrieved evidence, and the query.}
\label{fig:hierarchical_consistency_framework}\label{fig:hcf}
\end{figure*}

Figure~\ref{fig:hierarchical_consistency_framework} summarizes the
core contribution: three linked diagnostics localize degradation at
the knowledge corpus, retrieved evidence, and generated answer. The
horizontal arrows represent information flow through the RAG process;
the metric boxes report distinct diagnostic views rather than asserting
that one score determines another.

\subsection{Level 1 --- Corpus Consistency}\label{sec:intra}\label{sec:corpus}

Let $\mathcal{D} = \{d_1, \dots, d_N\}$ denote a corpus of $N$
documents. Corpus consistency is evaluated along two complementary
dimensions: intra-document metrics assess the structural quality of
individual documents, whereas inter-document metrics assess factual
agreement across documents. These dimensions are subsequently combined
into a unified Corpus Consistency metric.

\subsubsection{Intra-Document Metrics}\label{sec:intra_document_metrics}

\begin{definition}[Intra-Document Quality Score]
Let $Q:\mathcal{D}\rightarrow[0,1]$ be a document-quality estimator.
For $d\in\mathcal{D}$, define $\mathrm{IaDQ}(d)=Q(d)$ as the
normalized structural fitness of $d$ for retrieval, evaluated over
clarity, internal consistency, completeness, information density,
and coherence. IaDQ does not estimate factual truth.
\end{definition}

\begin{definition}[Intra-Document Normalized Integrity]
This metric is the mean normalized structural quality across the corpus:
\begin{equation}
  \mathrm{IaDNI} = \frac{1}{|\mathcal{D}|}
  \sum_{d \in \mathcal{D}} \mathrm{IaDQ}(d) \;\in [0,1].
\end{equation}
IaDNI equals one when every document has perfect structural quality
and approaches zero as average intra-document quality degrades.
\end{definition}

\subsubsection{Inter-Document Metrics}\label{sec:inter_document_metrics}

Inter-document consistency requires detecting factual divergences
across document boundaries. Because these conflicts are relational,
neither isolated document-quality assessment nor similarity-based
retrieval identifies them. Rather than relying on a pre-specified set
of tracked variables, we ground IeDCD in
\emph{atomic fact extraction}: the corpus itself generates the set
of trackable factual claims endogenously. This approach is directly
inspired by the atomic knowledge unit extraction framework of IndexRAG
\cite{bao2026indexrag} and the atomic fact evaluation methodology of
FActScore \cite{min2023factscore}.

The IeDCD pipeline proceeds in four steps, formalized below and
summarized in Algorithm~\ref{alg:iedcd}.

\begin{definition}[Atomic Fact Triple]
Let $\mathsf{Ent}$, $\mathsf{Rel}$, and $\mathsf{Val}$ denote the
spaces of entities, relations, and values, respectively. Define the
extraction operator
\begin{equation}
F:\mathcal{D}\rightarrow
\mathcal{P}(\mathsf{Ent}\times\mathsf{Rel}\times\mathsf{Val}),
\end{equation}
where $\mathcal{P}$ denotes the power set. For each
$d_i\in\mathcal{D}$, the corresponding source-linked atomic facts are
\begin{equation}
  \mathcal{A}_i = \bigl\{(d_i,e,r,v):(e,r,v)\in F(d_i)\bigr\},
\end{equation}
where $d_i$ is the source document identifier, $e$ is a named entity,
$r$ is a relation or attribute, and $v$ is the asserted value. Each
triple encodes one atomic, independently verifiable factual claim
extracted from $d_i$. For example, the sentence \emph{``The capital
of Spain is Madrid''} extracted from document $\texttt{doc}_1$ yields
the triple $(\texttt{doc}_1,\; \texttt{Spain},\; \texttt{capital},\;
\texttt{Madrid})$.
\end{definition}

\begin{definition}[Corpus Fact Index]
Let $\mathcal{A} = \bigcup_{i=1}^{N} \mathcal{A}_i$ be the union of
all atomic fact triples across the corpus. Prior to indexing, entity
and relation strings are normalized by a function
$\nu: \text{string} \to \text{string}$ applying lowercasing and
diacritic removal. Two values $v$ and $v'$ are treated as identical
if and only if $\nu(v) = \nu(v')$. The \emph{corpus fact index}
$\mathcal{M}$ is an in-memory map grouping triples by their normalized
$(e, r)$ key:
\begin{equation}
  \mathcal{M}(e, r) = \bigl\{(d_i, v) :
  (d_i, e, r, v) \in \mathcal{A} \bigr\}.
\end{equation}
Each entry collects all document--value pairs asserting a claim about
attribute $r$ of entity $e$ across the entire corpus. The reliability
of atomic fact extraction depends on the consistency of $F$.
Extraction coverage and semantic correspondence are assessed in Sections~\ref{sec:extraction} and~\ref{sec:correspondence}.
\end{definition}

\noindent\textit{Note on normalization.} The normalization strategy
$\nu$ is intentionally minimal to keep the framework tractable.
Richer strategies such as embedding-based entity clustering would
reduce missed conflicts from surface variation (e.g.,
\emph{Spain} vs.\ \emph{Kingdom of Spain}) but introduce additional
complexity. These are left for future work.

\begin{definition}[Inter-Document Conflict and Provenance]
An $(e, r)$ pair is \emph{conflicted} if the corpus fact index
contains at least two documents asserting different values:
\begin{equation}
\begin{aligned}
\mathrm{conflict}(e,r)=\mathbb{I}\!\bigl(&\exists (d_i,v),(d_j,v')
\in\mathcal{M}(e,r):\\
&d_i\neq d_j\ \wedge\ \nu(v)\neq\nu(v')\bigr).
\end{aligned}
\end{equation}
The \emph{conflict provenance set} is:
\begin{equation}
\begin{aligned}
\Pi(e,r)=\{d_i,d_j :\;&(d_i,v),(d_j,v')\in\mathcal{M}(e,r),\\
&d_i\neq d_j,\ \nu(v)\neq\nu(v')\}.
\end{aligned}
\end{equation}
$\Pi(e,r)$ identifies every document contributing to the conflict,
directly supporting the XAI contribution of this framework.
\end{definition}

\begin{definition}[Inter-Document Consistency Degree]
Let $\mathcal{K} = \{(e,r) : \mathcal{M}(e,r) \neq \emptyset\}$ be
the set of all $(e,r)$ pairs tracked in the corpus fact index. The
\emph{Inter-Document Consistency Degree} is:
\begin{equation}
  \mathrm{IeDCD} = 1 - \frac{
    \left|\bigl\{(e,r) \in \mathcal{K} :
    \mathrm{conflict}(e,r) = 1\bigr\}\right|
  }{|\mathcal{K}|} \;\in [0,1].
\end{equation}
$\mathrm{IeDCD} = 1$ when no $(e,r)$ pair carries conflicting
assignments. $\mathrm{IeDCD} = 0$ when every tracked pair is
contested. The provenance sets $\{\Pi(e,r)\}$ constitute an
\emph{explainability layer}: for any corpus with at least one detected
conflict, the practitioner can recover the exact
documents responsible for each conflict.
\end{definition}

\begin{algorithm}[t]
\caption{IeDCD Computation Pipeline}
\label{alg:iedcd}
\begin{algorithmic}[1]
\Require Corpus $\mathcal{D} = \{d_1, \dots, d_N\}$,
  fact extractor $F$
\Ensure $\mathrm{IeDCD} \in [0,1]$, provenance sets $\{\Pi(e,r)\}$

\Statex \textbf{Step 1: Atomic fact extraction}
\State $\mathcal{A} \leftarrow \emptyset$
\For{each $d_i \in \mathcal{D}$}
  \State $\mathcal{A}_i \leftarrow
    \{(d_i,e,r,v):(e,r,v)\in F(d_i)\}$
  \State $\mathcal{A} \leftarrow \mathcal{A} \cup \mathcal{A}_i$
\EndFor

\Statex \textbf{Step 2: Build corpus fact index}
\State $\mathcal{M} \leftarrow$ empty map
\For{each $(d_i, e, r, v) \in \mathcal{A}$}
  \State $\mathcal{M}(e, r) \leftarrow \mathcal{M}(e, r)
    \cup \{(d_i, v)\}$
\EndFor

\Statex \textbf{Step 3: Conflict detection}
\State $\mathcal{K} \leftarrow \{(e,r) :
  \mathcal{M}(e,r) \neq \emptyset\}$
\State $\mathcal{K}_\text{conflict} \leftarrow \emptyset$
\For{each $(e, r) \in \mathcal{K}$}
  \If{$\exists\, (d_i, v),\, (d_j, v') \in \mathcal{M}(e,r)$
    with $d_i \neq d_j$ and $\nu(v) \neq \nu(v')$}
    \State $\mathcal{K}_\text{conflict} \leftarrow
      \mathcal{K}_\text{conflict} \cup \{(e,r)\}$
    \State $\Pi(e,r) \leftarrow \{d_i, d_j :
      (d_i, v) \in \mathcal{M}(e,r),\;
      (d_j, v') \in \mathcal{M}(e,r),\;
      d_i \neq d_j,\; \nu(v) \neq \nu(v')\}$
  \EndIf
\EndFor

\Statex \textbf{Step 4: IeDCD aggregation}
\State \Return $\mathrm{IeDCD} = 1-|\mathcal{K}_\text{conflict}|\,
  /\, |\mathcal{K}|$,\quad
  $\{\Pi(e,r) \mid (e,r) \in \mathcal{K}_\text{conflict}\}$
\end{algorithmic}
\end{algorithm}

\subsubsection{Unified Corpus Consistency Metric}\label{sec:unified_corpus_consistency}

\begin{definition}[Corpus Consistency]
Corpus Consistency unifies the two consistency dimensions into a
single corpus-level consistency diagnostic:
\begin{equation}\label{eq:cc}
\begin{aligned}
\mathrm{CC}&=w_c\,\mathrm{IeDCD}+w_n\,\mathrm{IaDNI},\\
w_c,w_n&>0,\quad w_c+w_n=1,\quad \mathrm{CC}\in[0,1].
\end{aligned}
\end{equation}
$\mathrm{CC} = 1$ denotes a fully consistent corpus; $\mathrm{CC} = 0$
denotes a completely degraded one. The complementary degradation
score $\mathrm{CD}=1-\mathrm{CC}$ may be used where a higher-is-worse
orientation is analytically convenient.
\end{definition}

The linear form of CC follows the weighted aggregation model for
multi-dimensional data quality established by Pipino et al.\
\cite{pipino2002data}, where linearity ensures
\emph{decomposability}: the contribution of each quality dimension to
the overall score is independently interpretable and auditable. An
operator observing a low CC can immediately attribute it to low IeDCD,
low IaDNI, or both, in proportion to their weighted contributions. The
weights $w_c$ and $w_n$ are practitioner-configurable: a
regulated domain may set $w_c > w_n$ to penalize factual
inconsistency more heavily; a corpus with known structural quality
issues may set $w_n > w_c$.

In this paper we set $w_c = w_n = 0.5$, treating both quality
dimensions as equally important in the absence of domain-specific
prior knowledge. Future work should examine sensitivity to these
weights and alternative aggregation functions, including nonlinear
forms that capture interactions between factual conflict and structural
noise while preserving boundedness, monotonicity, and interpretability
(Section~\ref{sec:limits}).

\subsection{Level 2: Retrieval Consistency}
\label{subsec:retrieval_consistency}\label{sec:retrieval}

Retrieval Consistency characterizes the relevance and score dispersion
of the evidence selected for a query. We quantify it using three
indicators derived from query--chunk similarity scores.

Let:
\begin{equation}
\mathcal{R}(q) = \{c_1, c_2, \dots, c_k\},\qquad
k=|\mathcal{R}(q)|,
\end{equation}

denote the set of retrieved chunks for query $q$, where $k$ is the
number of retrieved chunks. Let
$\mathrm{sim}$ be the normalized semantic similarity function used by
the retriever; it maps a query--chunk pair to $[0,1]$. The score of
chunk $c_i$ is
\begin{equation}
s_i(q) = \mathrm{sim}(q,c_i),\qquad
\mathrm{sim}:\mathrm{text}\times\mathrm{text}\rightarrow[0,1].
\end{equation}

We define the following retrieval consistency indicators:

\paragraph{Top Retrieval Similarity ($SS_1(q)$)}
Measures the similarity of the highest-ranked retrieved chunk:
\begin{equation}
SS_1(q) = \max_{1\leq i\leq k} s_i(q).
\end{equation}

This metric captures the strength of the single most relevant retrieved evidence source.

\paragraph{Mean Retrieval Similarity ($SS_\mu(q)$)}
Measures the average semantic agreement across the retrieved evidence set:
\begin{equation}
SS_\mu(q) = \frac{1}{k} \sum_{i=1}^{k} s_i(q).
\end{equation}

Higher values indicate stronger overall retrieval coherence.

\paragraph{Retrieval Similarity Variance ($SS_\sigma(q)$)}
Measures the dispersion of similarity scores:
\begin{equation}
SS_\sigma(q) = \frac{1}{k}\sum_{i=1}^{k}
\bigl(s_i(q)-SS_\mu(q)\bigr)^2 \in [0,1/4].
\end{equation}

Because every $s_i(q)$ lies in $[0,1]$, the variance is at most
$(1-0)^2/4=1/4$, independently of $k$. This quantity is dispersion, not consistency by
itself. High values
indicate heterogeneous relevance scores; unusually low values can
also be diagnostic when contradictory but topically similar chunks
are compressed into the same embedding neighborhood. Its direction
must therefore be interpreted jointly with CC and the retrieved text.

\subsection{Level 3: Answer Consistency}
\label{subsec:answer_consistency}\label{sec:answer}

The final level of the framework evaluates whether the generated answer remains grounded in the retrieved evidence and aligned with the original query context. We define \emph{Answer Consistency} as the degree of agreement between:
(i) the generated answer,
(ii) the retrieved evidence,
and (iii) the user query.

Unlike Retrieval Consistency, Answer Consistency evaluates the
synthesis stage. Let $q$ be a query, $\mathcal{R}(q)$ its retrieved
evidence, $a$ the generated answer, and
$J(q,a,\mathcal{R}(q))\in[0,1]$ an evaluator of query alignment,
evidence support, and absence of contradiction. We define
\begin{equation}
  \mathrm{ACS}(q,a)=J\bigl(q,a,\mathcal{R}(q)\bigr).
\end{equation}
The evaluator $J$ may be rule-based, learned, or hybrid; the
experimental instantiation is specified in
Section~\ref{sec:measurement_protocol}.

Importantly, ACS is not intended as a probabilistic confidence estimate or a measure of factual truthfulness. A highly consistent answer may still be factually incorrect if the retrieved evidence itself is incorrect. Instead, ACS measures grounded agreement within the RAG pipeline.

\section{Experimental Design}\label{sec:experimental_design}\label{sec:design}
This section describes the experimental platform and estimator
instantiation, the controlled corpora, the rationale for their design,
the corpus generation and review process, the query set, and the human
reference-answer assessment. The numerical query records used in the
analysis are reported in the appendices.

\subsection{Experimental Platform}

The experiments were executed in Mentomy, which provided knowledge-base
ingestion, retrieval, role-based access control, answer generation, and
evaluation workflows \cite{mentomy2026platform}. Mentomy is the
experimental platform only; the HCF definitions and operators are
platform-agnostic.

\subsection{Operator and Retrieval Configuration}
\label{sec:measurement_protocol}

The generic operators $Q$, $F$, and $J$ defined in
Section~\ref{sec:problem_statement} were implemented through
three task-specific instruction sets supplied to Mistral Small~4.
LLM-based assessment is increasingly used for scalable NLP and RAG evaluation
\cite{zheng2023judging,es2023ragas}, although its
reliability depends on the task, scoring rubric, and evaluator. Known
risks include position, verbosity, and self-enhancement biases
\cite{zheng2023judging}. LLM instantiation is an experimental choice,
not a requirement of HCF, and its recorded explanations are analyzed separately from the author's reference-answer assessment.

The document-quality assessor $Q$ scored clarity, internal consistency,
completeness, information density, and coherence without external
fact-checking. The atomic-fact extractor $F$ returned source-linked
$(e,r,v)$ claims; documents were processed in overlapping windows with
per-window and per-document extraction caps to bound latency. The
answer-consistency judge $J$ assessed query alignment, support from the
retrieved evidence, and absence of unsupported or contradictory
synthesis, returning ACS in $[0,1]$ together with a textual
justification.

All three operators and answer generation used Mistral Small~4
(\texttt{mistral-small-2603}), released on 16 March 2026. The model has
119 billion total parameters and was run with temperature $0.1$
\cite{mistral2026small4}. Separate task instructions isolated the roles
of document-quality assessment, fact extraction, answer generation,
and answer-consistency evaluation.

Retrieval used a Pinecone dense-vector index with cosine similarity
\cite{pinecone2026index}. Text was embedded with the Microsoft-developed
\texttt{multilingual-e5-large} model, hosted through Pinecone Inference
\cite{wang2024multilinguale5,pinecone2026e5}. The index dimension was
1{,}024; the model produces dense text vectors and accepts at most 507
input tokens per sequence. For each query, semantic search selected 15
candidates, the multilingual \texttt{BAAI/bge-reranker-v2-m3}
cross-encoder reranked the query--passage pairs and retained 10
\cite{nogueira2019passage,chen2024bgereranker}, and
role-based access control removed
chunks unavailable to the requesting user. The final authorized set is
$\mathcal{R}(q)$, so $k(q)=|\mathcal{R}(q)|$ varies across queries.
$SS_1$, $SS_\mu$, and $SS_\sigma$ were computed over this final set.

\subsection{Controlled Corpus Construction}

We construct four controlled corpora across five domains --- physics,
medicine, history, geography, and biology. Table~\ref{tab:corpora}
summarizes the four conditions.

\begin{table*}[!t]
\centering
\small
\begin{tabular}{lccp{5cm}}
\toprule
Corpus & Docs & Words & Composition \\
\midrule
A (Baseline)  & 5  & 12{,}659 & 5 clean documents \\
B (Conflict)  & 10 & 25{,}274 & 5 clean + 5 conflict-injected \\
C (Noise)     & 5  & 13{,}327 & 5 structurally degraded \\
D (Combined)  & 20 & 49{,}548 & 5 clean + 5 conflict-injected + 5 degraded
                                  + 5 combined \\
\bottomrule
\end{tabular}
\caption{Four controlled corpora with isolated degradation modes across five science and humanities domains (physics, medicine, history, geography, biology).}
\label{tab:corpora}
\end{table*}

\textbf{Corpus A --- Clean baseline (5 documents).} One document per
domain, approximately 2{,}000--2{,}800 words each. Every document is
internally consistent, well-structured, and contains no factual
conflicts with any other document in the corpus.

\textbf{Corpus B --- Low IeDCD (10 documents).} The 5 documents from
Corpus A plus 5 conflict-injected counterparts, one per domain. Correct
factual values are replaced with plausible but incorrect alternatives
of the same type and order of magnitude, preserving document fluency
and topical relevance. Representative examples include the speed of
light stated as 250{,}000~km/s instead of 299{,}792~km/s, high-intensity
atorvastatin dosed at up to 160~mg daily instead of 80~mg, and the
execution of Louis~XVI dated to 28 January 1793 instead of 21 January
1793.

\textbf{Corpus C --- Low IaDNI (5 documents).} The 5 documents from
Corpus A with structural degradation applied. Six degradation operators
are used --- truncation, formatting removal, irrelevant content
injection, paragraph repetition, inconsistent terminology, and broken
cross-references --- with at least three operators applied per document.
No factual conflicts are introduced.

\textbf{Corpus D --- Combined degradation (20 documents).} A union of
all degradation conditions: 5 clean documents, 5 conflict-injected
documents, 5 structurally degraded documents, and 5 newly constructed
documents that combine conflict injection and structural degradation.
This condition represents unmanaged corpus growth in which clean source
material, contradictory copies, degraded fragments, and fully corrupted
documents coexist within one retrievable index.

\subsection{Rationale for a Controlled Corpus}
We construct controlled corpora rather than evaluating an organic
collection to obtain ground truth and isolate the degradation modes.

\textbf{Ground-truth availability.} With a controlled corpus we know
exhaustively which $(e,r)$ pairs are in conflict, which documents assert
each value, and what the canonical correct value is. The complete
provenance map $\Pi(e,r)$ is maintained during the experiment, permitting
source-linked inspection of extracted conflicts. For organic corpora, complete factual ground truth is in
general unavailable.

\textbf{Isolability.} The four corpora form a $2\times 2$ factorial
design over \{low, high\}~$\times$~\{low, high\} levels of IeDCD and
IaDNI, with Corpus~D additionally varying corpus size. In organic
collections these two consistency dimensions may covary, preventing
clean attribution of observed ACS degradation to a single
cause. The controlled design eliminates this confounding by
construction.

\subsection{Corpus Generation and Review Process}

To maximize the factual accuracy and linguistic quality of the
baseline documents (Corpus A), we adopted a multi-model peer review
pipeline inspired by ensemble validation practices in data-centric AI
\cite{zha2023datacentric}. Each document was generated with Anthropic
Claude Opus~4.7 (\texttt{claude-opus-4-7})
\cite{anthropic2026models}. The generated documents were then reviewed
by OpenAI GPT-5.5 (\texttt{gpt-5.5}) and Google Gemini~3.5 Flash
\cite{openai2026models,google2026gemini}, each prompted to identify factual errors, internal
inconsistencies, unclear passages, and deviations from canonical
sources. Corrections flagged by either reviewer were applied before
the document was admitted to Corpus A. This review process was intended
to reduce factual errors, internal inconsistencies, and structural
defects before inclusion.

\subsection{Query Set Design}
We evaluate 100 query--corpus instances, obtained by issuing the same
25 queries to each of the four corpus conditions. The fixed query set enables
within-query paired comparison across degradation conditions: every
observed change in ACS for a given query is attributable to the corpus
condition rather than to query variation.

Queries are stratified across three types. Q1--Q10 target manipulated
entity--relation values, Q11--Q20 serve as controls, and Q21--Q25 require
multi-part synthesis. These labels describe the intended probes; the final
retrieved context remains determined by retrieval, reranking, and authorization.

\subsection{Human Reference-Answer Assessment}

One human evaluator manually compared every generated response with the
supplied ground-truth response, yielding 100 response--reference
comparisons. Exact matches and semantically equivalent expressions,
including compatible translations, units, and numerical precision, were
accepted; cases with missing or materially different content were retained
as such in the result register. This assessment establishes whether a
response matches the benchmark reference. It is separate from ACS, which
assesses agreement with the retrieved context, and does not constitute an
independent validation of the ACS evaluator.

\section{Results}\label{sec:results}
This section shows the results and evaluates the implications of the proposed framework through ingestion diagnostics, extraction recovery, cross-document correspondence, and query-level answer consistency.

\subsection{Ingestion-Phase Corpus Diagnostics}\label{sec:ingestion}
The goal is to determine whether the component diagnostics distinguish the intended corpus conditions. Table~\ref{tab:ingestion} reports the supplied ingestion measurements and CC calculated from the reported component scores using~\eqref{eq:cc}.

\begin{table*}[!t]\centering\small
\caption{Ingestion diagnostics. Higher IaDNI, IeDCD, and CC indicate better estimated consistency. ER pairs and conflicts are reported index counts.}\label{tab:ingestion}
\begin{tabular}{lrrrrr}\toprule
Corpus & IaDNI & IeDCD & ER pairs & Conflicts & CC\\\midrule
A (baseline) & 0.82 & 1.00 & 545 & 0 & 0.910\\
B (conflict) & 0.81 & 0.93 & 882 & 58 & 0.870\\
C (noise) & 0.23 & 1.00 & 489 & 0 & 0.615\\
D (combined) & 0.55 & 0.89 & 1,438 & 148 & 0.720\\\bottomrule
\end{tabular}
\end{table*}

A and B have similar integrity (0.82 and 0.81), while only B has flagged divergences. This is consistent with modifications that preserve structure while altering values. C instead combines low integrity (0.23) with no detected cross-document divergences. Its CC of 0.615 emphasizes why an absence of detected conflict does not establish a usable knowledge base. D combines intermediate integrity with the largest flagged-key count and a reported CC of 0.720. Its position above C reflects the mixture of clean and modified documents, not a failure of arithmetic or a requirement that the combined condition rank worst. The absolute number of flags also depends on the number and coverage of extracted keys.

\subsection{Evaluation of Atomic Fact Extraction}\label{sec:extraction}
The goal is to assess the extraction stage of Algorithm~\ref{alg:iedcd} against 40 selected reference facts in Corpus A. Recovery is judged semantically: language changes and equivalent expressions count as recovered, including records that preserve the target fact without reproducing the full wording. Table~\ref{tab:recovery} retains this assessment from the ingestion analysis.

\begin{table*}[!t]\centering\small
\caption{Recovery of selected reference facts in Corpus A.}\label{tab:recovery}
\begin{tabular}{lrrrr}\toprule
Domain & Reference & Recovered & Missed & Rate\\\midrule
Physics & 6 & 5 & 1 & 83.3\%\\
Medicine & 8 & 7 & 1 & 87.5\%\\
History & 9 & 5 & 4 & 55.6\%\\
Geography & 9 & 9 & 0 & 100.0\%\\
Biology & 8 & 8 & 0 & 100.0\%\\\midrule
Total & 40 & 34 & 6 & 85.0\%\\\bottomrule
\end{tabular}\end{table*}

The extractor recovers 34 of 40 facts (85.0\%). Geography and biology have complete recovery within this selected set, while history accounts for four of the six omissions. Thus the extraction coverage is useful but uneven across domains. Table~\ref{tab:examples} illustrates direct recoveries and a missing authorship relation.

\begin{table*}[!t]\centering\small
\caption{Examples from the Corpus A extraction analysis.}\label{tab:examples}
\begin{tabular}{l>{\raggedright\arraybackslash}p{5.1cm}>{\raggedright\arraybackslash}p{7.4cm}l}\toprule
Domain & Reference fact & Extracted entity--relation--value & Outcome\\\midrule
Physics & Speed of light: 299,792,458 m/s & (speed of light in vacuum, exact value, 299,792,458 m/s) & Recovered\\
Medicine & High-intensity atorvastatin: 40--80 mg/day & (atorvastatin, high dose, 40--80 mg/day) & Recovered\\
Geography & Aconcagua elevation: 6,961 m & (Aconcagua, elevation, 6,961 m) & Recovered\\
History & Edmund Burke authored \emph{Reflections on the Revolution in France} & No corresponding authorship record & Missed\\\bottomrule
\end{tabular}\end{table*}

\subsection{Evaluation of ER-Pair Correspondence}\label{sec:correspondence}
The goal is to inspect whether extracted clean-side and conflict-injected records can be recognized as the same entity--relation pair before comparing values. This analysis directly compares selected extracted records associated with manipulated relations; it does not repeat the preceding reference-fact evaluation. Equivalent language, dates, units, and minor relation labels are accepted in the semantic review. In particular, ``exact value'' and ``precise value'' are treated as corresponding relations, while different factual values remain available for comparison.

\begin{table*}[!t]\centering\small\setlength{\tabcolsep}{3pt}
\caption{Semantic correspondence among selected Corpus B records. Rate is matched divided by clean-side records.}\label{tab:correspondence}
\begin{tabular}{lrrrr}\toprule
Domain & Clean & Modified & Matched & Rate\\\midrule
Physics & 6 & 6 & 5 & 83.3\%\\
Medicine & 7 & 7 & 6 & 85.7\%\\
History & 5 & 6 & 5 & 100.0\%\\
Geography & 2 & 6 & 2 & 100.0\%\\
Biology & 7 & 6 & 6 & 85.7\%\\\midrule
Total & 27 & 31 & 24 & 88.9\%\\\bottomrule
\end{tabular}\end{table*}

Table~\ref{tab:correspondence} shows that 24 of the 27 selected
clean-side entity--relation records were matched to a corresponding
record in a modified document, for an overall correspondence rate of
88.9\%. History and geography achieved complete correspondence within
their selected records, while physics, medicine, and biology each had
one unmatched clean-side record. Table~\ref{tab:match_examples} then
shows how successful correspondences expose conflicting values and how
an absent counterpart appears in the analysis.

\begin{table*}[!t]\centering\small
\caption{Corpus B correspondence examples, translated into English.}\label{tab:match_examples}
\begin{tabular}{l>{\raggedright\arraybackslash}p{5cm}>{\raggedright\arraybackslash}p{5cm}>{\raggedright\arraybackslash}p{3.3cm}}\toprule
Domain & Clean extracted record & Modified extracted record & Outcome\\\midrule
Medicine & (atorvastatin, high dose, 40--80 mg/day) & (atorvastatin, high dose, 40--160 mg/day) & Successfully matched; conflicting values\\
Physics & (Planck constant, approximate value, $6.626\times10^{-34}$ J$\cdot$s) & (Planck constant, approximate value, $6.200\times10^{-34}$ J$\cdot$s) & Successfully matched; conflicting values\\
Medicine & (statins, downstream effect, increased LDL receptors) & No corresponding modified record & Missing counterpart\\\bottomrule
\end{tabular}\end{table*}

\subsection{Query-Level Answer and Retrieval Diagnostics}\label{sec:queryresults}
The goal is to evaluate answer consistency and its relationship with
retrieval relevance. The human evaluator completed all 100
response--ground-truth comparisons before this analysis. Those records
establish answer--reference agreement; ACS addresses a different
question---whether the answer is supported without contradiction by the
final authorized context.

\begin{figure*}[!t]\centering
\includegraphics[width=.98\textwidth]{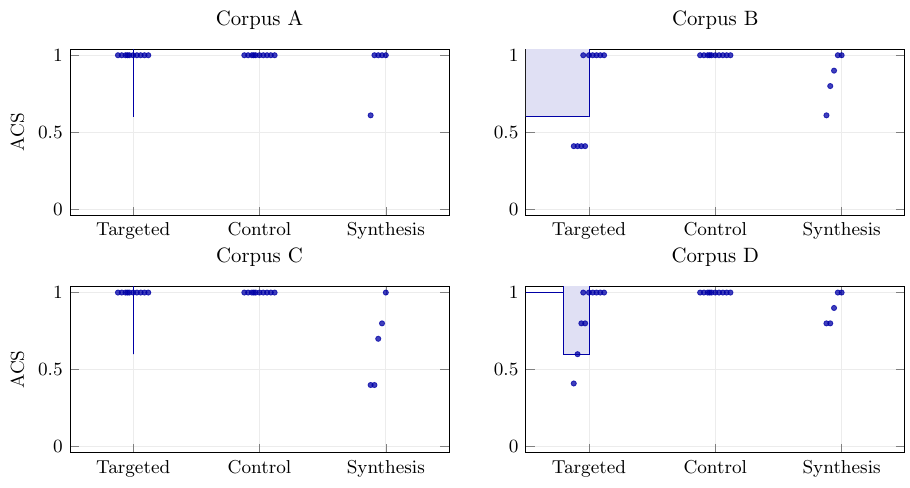}
\caption{ACS distributions by question type and corpus. Targeted questions (Q1--Q10) address manipulated relations, control questions (Q11--Q20) address unmodified facts, and synthesis questions (Q21--Q25) combine multiple facts. Points show individual questions; grouping avoids implying an ordered trend across nominal question identifiers.}
\label{fig:queryresults}
\vspace{0.6em}
\includegraphics[width=.98\textwidth]{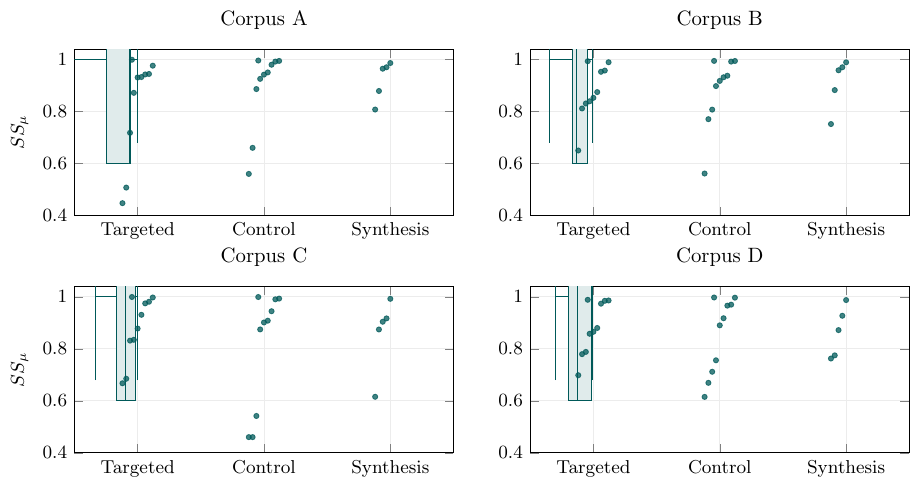}
\caption{Distributions of mean retrieval similarity $SS_\mu$ by question type and corpus. Targeted questions (Q1--Q10) address manipulated relations, control questions (Q11--Q20) address unmodified facts, and synthesis questions (Q21--Q25) combine multiple facts. Points show individual questions. High semantic relevance can coexist with reduced ACS because similarity does not test factual compatibility.}
\label{fig:retrievalresults}
\end{figure*}

Figure~\ref{fig:queryresults} first presents the ACS distributions for
the targeted, control, and synthesis questions in each corpus.

Figure~\ref{fig:queryresults} shows that the median ACS is 1.00 in every
corpus, but the distributions differ by question type. Corpus B has the
largest reduction among targeted questions, which are designed to retrieve
the manipulated relations. Corpus C retains ACS 1.00 for targeted and
control questions but shows lower values for several synthesis questions.
Corpus D shows reductions in both targeted and synthesis questions. These
patterns demonstrate that a corpus-level score does not determine the
consistency of an individual answer; the final evidence retrieved for each
query remains decisive.

Figure~\ref{fig:retrievalresults} presents the corresponding $SS_\mu$
distributions. Values remain high across most groups, including Corpus B's
targeted questions, in contrast to the lower targeted-question ACS
distribution in Figure~\ref{fig:queryresults}. In aggregate, B has the
highest mean $SS_\mu$ (0.8843) but the lowest mean ACS (0.8780). Retrieved
chunks can therefore remain strongly related to the question while
disagreeing about the requested fact; semantic relevance and evidence
consistency are not the same property.

Table~\ref{tab:query_summary} next consolidates the 25 records for each
corpus as arithmetic means and sample standard deviations. The standard
deviations describe variation across the fixed query set, not uncertainty
over repeated experimental runs.

\begin{table*}[!t]
\centering\small\setlength{\tabcolsep}{5pt}
\caption{Corpus-level summaries of the 25 query records reported individually in Tables~\ref{tab:queryA}--\ref{tab:queryD}. CC is fixed within a corpus; all remaining entries are mean $\pm$ sample standard deviation across queries.}
\label{tab:query_summary}
\begin{tabular}{lcccccc}
\toprule
Corpus & CC & $k(q)$ & $SS_1$ & $SS_\mu$ & $SS_\sigma$ & ACS\\
& & \multicolumn{5}{c}{mean $\pm$ standard deviation}\\
\midrule
A & 0.910 & $1.40\pm0.50$ & $0.9371\pm0.0941$ & $0.8705\pm0.1619$ & $0.0223\pm0.0537$ & $0.9844\pm0.0780$\\
B & 0.870 & $2.60\pm0.71$ & $0.9431\pm0.0929$ & $0.8843\pm0.1129$ & $0.0168\pm0.0398$ & $0.8780\pm0.2257$\\
C & 0.615 & $1.28\pm0.46$ & $0.8928\pm0.1153$ & $0.8462\pm0.1713$ & $0.0163\pm0.0464$ & $0.9320\pm0.1749$\\
D & 0.720 & $4.68\pm0.80$ & $0.9605\pm0.0494$ & $0.8646\pm0.1173$ & $0.0257\pm0.0388$ & $0.9244\pm0.1493$\\
\bottomrule
\end{tabular}
\end{table*}

Table~\ref{tab:query_summary} shows that mean ACS is 0.9844, 0.8780,
0.9320, and 0.9244 for Corpora A--D, respectively. Corpus B has the
lowest mean ACS even though its CC exceeds those of C and D, confirming
that the three HCF levels provide complementary rather than interchangeable
diagnostics. The retrieval variables also require distinct interpretations.
$SS_1$ measures whether at least one highly relevant chunk survives the
complete retrieval and authorization workflow. Its means remain high
($0.8928$--$0.9605$), so a strong top hit does not establish that the
remaining context is compatible. $SS_\mu$ measures average relevance over
all final authorized chunks and varies only from 0.8462 to 0.8843.
$SS_\sigma$ measures within-query score dispersion; it is not inherently
better when high or low, is zero whenever only one chunk remains, and must
therefore be interpreted together with $k(q)$.

Finally, Table~\ref{tab:acs_examples} presents four Corpus B cases in
which the human evaluator accepted the answer as matching the ground truth,
while ACS identified an incompatible statement in the retrieved context.

\begin{table*}[!t]\centering\small\setlength{\tabcolsep}{4pt}
\caption{Reference-matching answers with contradictory retrieved evidence in Corpus B.}\label{tab:acs_examples}
\begin{tabular}{>{\raggedright\arraybackslash}p{1.15cm}>{\raggedright\arraybackslash}p{3.0cm}>{\raggedright\arraybackslash}p{2.8cm}>{\centering\arraybackslash}p{1.15cm}>{\raggedright\arraybackslash}p{6.8cm}}\toprule
Case & Reference-matching answer & Conflicting statement identified & ACS (score) & ACS (explanation)\\\midrule
B--Q1 & Speed of light in vacuum: 299,792,458 m/s & Approximately 250,000 km/s & 0.41 & ``The answer agrees with the exact value in the clean passage, but another passage contradicts it with the lower value.''\\\addlinespace
B--Q4 & Statin therapy is typically indefinite & Discontinuation after one to three years if lipid targets are achieved & 0.41 & ``The answer states indefinite treatment, which conflicts with contextual guidance allowing discontinuation after one to three years.''\\\addlinespace
B--Q6 & Louis XVI was executed on 21 January 1793 & Execution on 28 January 1793 & 0.41 & ``The answer's date agrees with the presented evidence, but another part of the context gives a contradictory date.''\\\addlinespace
B--Q9 & Aconcagua's elevation is 6,961 m & Elevation of 6,400 m & 0.41 & ``The answer is supported by the document, but another part of the text gives 6,400 m, contradicting that elevation.''\\\bottomrule
\end{tabular}
\end{table*}

Table~\ref{tab:acs_examples} demonstrates the principal answer-level
contribution. In every case, the answer selects the ground-truth value, but
the final context also contains a conflicting value or date. ACS therefore
does not simply repeat the human correctness judgment: it reports that the
answer is supported and simultaneously identifies the statement that makes
its evidence contradictory. The accompanying explanation makes the reduced
score actionable by directing the reader to the competing claim.

\section{Discussion}\label{sec:discussion}
This section interprets the principal findings and the scope of the resulting audit.

The central strength is the distinction between an acceptable answer and its evidential setting. In Table~\ref{tab:acs_examples}, an answer-only check would accept each response, whereas ACS also identifies an incompatible retrieved statement and explains the disagreement. This is actionable diagnostic information: a reviewer can inspect the competing source rather than simply replace a correct answer. The contribution is therefore not merely a lower score under degradation, but an explanation of evidence inconsistency that would remain invisible in reference-only correctness assessment.

The second strength is staged inspection. Corpus diagnostics separate estimated structural integrity from indexed divergence, while the extraction and correspondence analyses reveal whether records are available for comparison. This organization makes omissions and representation differences visible rather than attributing every failed comparison to the answer generator.

The third finding concerns the limits of similarity. High and nearly equal relevance scores can coexist with conflicting evidence, as illustrated by B--Q1 in Table~\ref{tab:acs_examples}. Relevance statistics remain useful for describing what was retrieved, but should not be labeled semantic coherence. A low variance can also result trivially from retaining one chunk. Corpus composition, ranking, and authorization jointly determine which evidence reaches the answer stage, so no monotonic relationship among CC, variance, and ACS is assumed.

Operationally, a flagged case should lead to source inspection, not an automatic declaration that the answer is false. Candidate divergence records identify documents to review; answer-level explanations identify statements or gaps to check. Threshold-based actions, such as abstention or source repair, would require separately validated decision rules. The experiments demonstrate the information that HCF exposes and how it localizes a problem, but they do not establish the false-positive or false-negative rate of a particular ACS cutoff because no independently labeled calibration set was used.

\section{Conclusions}\label{sec:conclusion}
HCF organizes RAG auditing across corpus integrity and factual divergence, final-context retrieval statistics, and answer--context consistency. Its model-independent formulation and source-linked records provide a common structure for examining where evidence limitations arise. The controlled experiments demonstrate useful extraction coverage and semantic correspondence, and illustrate the central distinction that a correct answer can coexist with contradictory retrieved evidence. The results also show that similarity dispersion and corpus-level consistency do not determine answer consistency in these four conditions. HCF is consequently best understood as a complementary diagnostic framework, not a replacement for factual verification or an independently validated probability of correctness.

\section{Limitations and Future Research}\label{sec:limits}
This section defines the boundaries of the evidence and the main priorities for further validation. The experiments cover 40 document placements across four configurations, representing 20 underlying document variants in five domains. Controlled modifications facilitate inspection, but the small, constructed collections cannot establish generalization to organic enterprise knowledge bases. The conditions differ in size and redundancy, and there is only one retained run per question and condition.

The single human evaluator's comparison of all 100 responses with their
ground-truth responses supports the reported benchmark-match assessments,
but it is not an independent validation of ACS. An independently adjudicated set of
answer--context pairs, including compatible paraphrases, unsupported claims,
and genuine contradictions, is needed to evaluate ACS sensitivity,
specificity, stability, and calibration. Reusing a model family for
generation and evaluation may introduce shared biases.

Algorithm~\ref{alg:iedcd} depends on extraction coverage and key normalization. Equivalent multilingual expressions, legitimately multi-valued relations, and omitted temporal qualifiers can distort its candidate-divergence counts. The semantic correspondence review is broader than the minimal runtime comparator and must not be presented as proof of automated semantic matching.

Future controlled ablations should vary the operator prompts, score
provenance, chunking parameters, and authorization configuration independently, repeat runs, and
compare HCF with established reference-free evaluation metrics. A larger
set of corpora distributed along controlled consistency gradients would
also be required to test whether upstream degradation statistically
propagates to downstream scores; the present four conditions support
decomposition and localization, not that conjecture.

\section*{Data and Code Availability}
The dataset release described in Appendix~\ref{app:dataset} contains the
documents forming Corpora A--D, the 25 queries, and their ground-truth
responses. The Mentomy platform is not included. Extracted records and
answer excerpts in the main text are translated or summarized for
readability.

\begingroup\emergencystretch=1em
\endgroup

\appendices
\section{Dataset Documentation}\label{app:dataset}\label{app:queries}
The public dataset is available at \url{https://github.com/ramongrobot/mentomy}. It contains the documents forming Corpora A--D, the 25 evaluation questions, and their supplied ground-truth responses. It does not contain Mentomy's proprietary platform implementation. The questions are organized as follows: Q1--Q10 target manipulated relations, Q11--Q20 are controls, and Q21--Q25 require synthesis.
\newpage
\raggedbottom
\noindent\begin{minipage}{\columnwidth}
\section{Corpus A Query Results}\label{app:A}
\begin{table}[H]\centering\footnotesize\setlength{\tabcolsep}{4pt}
\caption{Corpus A query-level results.}\label{tab:queryA}
\begin{tabular}{rrrrrr}\toprule Q & $k(q)$ & $SS_1$ & $SS_\mu$ & $SS_\sigma$ & ACS\\\midrule
1 & 2 & 0.9915 & 0.9443 & 0.0022 & 1.00\\
2 & 1 & 0.9990 & 0.9990 & 0.0000 & 1.00\\
3 & 1 & 0.9308 & 0.9308 & 0.0000 & 1.00\\
4 & 1 & 0.8717 & 0.8717 & 0.0000 & 1.00\\
5 & 2 & 0.9878 & 0.9424 & 0.0021 & 1.00\\
6 & 2 & 0.9922 & 0.9761 & 0.0003 & 1.00\\
7 & 1 & 0.9325 & 0.9325 & 0.0000 & 1.00\\
8 & 2 & 0.9986 & 0.7180 & 0.0787 & 1.00\\
9 & 2 & 0.8507 & 0.4475 & 0.1626 & 1.00\\
10 & 2 & 0.9453 & 0.5074 & 0.1917 & 1.00\\
11 & 1 & 0.9798 & 0.9798 & 0.0000 & 1.00\\
12 & 1 & 0.9255 & 0.9255 & 0.0000 & 1.00\\
13 & 2 & 0.9914 & 0.6602 & 0.1097 & 1.00\\
14 & 1 & 0.9418 & 0.9418 & 0.0000 & 1.00\\
15 & 2 & 0.9828 & 0.8862 & 0.0093 & 1.00\\
16 & 1 & 0.9920 & 0.9920 & 0.0000 & 1.00\\
17 & 1 & 0.9960 & 0.9960 & 0.0000 & 1.00\\
18 & 2 & 0.9979 & 0.9945 & 0.0000 & 1.00\\
19 & 1 & 0.9502 & 0.9502 & 0.0000 & 1.00\\
20 & 1 & 0.5600 & 0.5600 & 0.0000 & 1.00\\
21 & 2 & 0.9890 & 0.9863 & 0.0000 & 1.00\\
22 & 1 & 0.9701 & 0.9701 & 0.0000 & 1.00\\
23 & 1 & 0.8075 & 0.8075 & 0.0000 & 0.61\\
24 & 1 & 0.8788 & 0.8788 & 0.0000 & 1.00\\
25 & 1 & 0.9645 & 0.9645 & 0.0000 & 1.00\\
\bottomrule\end{tabular}
\end{table}\end{minipage}\par\medskip
\noindent\begin{minipage}{\columnwidth}
\section{Corpus B Query Results}\label{app:B}
\begin{table}[H]\centering\footnotesize\setlength{\tabcolsep}{4pt}
\caption{Corpus B query-level results.}\label{tab:queryB}
\begin{tabular}{rrrrrr}\toprule Q & $k(q)$ & $SS_1$ & $SS_\mu$ & $SS_\sigma$ & ACS\\\midrule
1 & 3 & 0.9914 & 0.9574 & 0.0019 & 0.41\\
2 & 2 & 0.9990 & 0.9934 & 0.0000 & 1.00\\
3 & 3 & 0.9308 & 0.8526 & 0.0032 & 1.00\\
4 & 3 & 0.9949 & 0.8746 & 0.0094 & 0.41\\
5 & 3 & 0.9878 & 0.9527 & 0.0016 & 1.00\\
6 & 2 & 0.9922 & 0.9894 & 0.0000 & 0.41\\
7 & 4 & 0.9417 & 0.8308 & 0.0121 & 1.00\\
8 & 3 & 0.9991 & 0.8116 & 0.0701 & 1.00\\
9 & 2 & 0.8507 & 0.8393 & 0.0001 & 0.41\\
10 & 3 & 0.9453 & 0.6500 & 0.1685 & 1.00\\
11 & 3 & 0.9796 & 0.8072 & 0.0435 & 1.00\\
12 & 2 & 0.9263 & 0.8977 & 0.0008 & 1.00\\
13 & 3 & 0.9920 & 0.7705 & 0.0979 & 1.00\\
14 & 2 & 0.9418 & 0.9316 & 0.0001 & 1.00\\
15 & 3 & 0.9828 & 0.9176 & 0.0083 & 1.00\\
16 & 2 & 0.9919 & 0.9918 & 0.0000 & 1.00\\
17 & 3 & 0.9960 & 0.9945 & 0.0000 & 1.00\\
18 & 4 & 0.9979 & 0.9942 & 0.0000 & 1.00\\
19 & 2 & 0.9504 & 0.9375 & 0.0002 & 1.00\\
20 & 2 & 0.5643 & 0.5615 & 0.0000 & 1.00\\
21 & 3 & 0.9947 & 0.9891 & 0.0000 & 0.90\\
22 & 1 & 0.9701 & 0.9701 & 0.0000 & 0.80\\
23 & 3 & 0.8073 & 0.7517 & 0.0023 & 0.61\\
24 & 2 & 0.8858 & 0.8823 & 0.0000 & 1.00\\
25 & 2 & 0.9645 & 0.9585 & 0.0000 & 1.00\\
\bottomrule\end{tabular}
\end{table}\end{minipage}\par\medskip
\noindent\begin{minipage}{\columnwidth}
\section{Corpus C Query Results}\label{app:C}
\begin{table}[H]\centering\footnotesize\setlength{\tabcolsep}{4pt}
\caption{Corpus C query-level results.}\label{tab:queryC}
\begin{tabular}{rrrrrr}\toprule Q & $k(q)$ & $SS_1$ & $SS_\mu$ & $SS_\sigma$ & ACS\\\midrule
1 & 1 & 0.9973 & 0.9973 & 0.0000 & 1.00\\
2 & 2 & 0.9953 & 0.9809 & 0.0002 & 1.00\\
3 & 1 & 0.8779 & 0.8779 & 0.0000 & 1.00\\
4 & 1 & 0.6674 & 0.6674 & 0.0000 & 1.00\\
5 & 2 & 0.9561 & 0.9307 & 0.0006 & 1.00\\
6 & 2 & 0.9908 & 0.9746 & 0.0003 & 1.00\\
7 & 1 & 0.6848 & 0.6848 & 0.0000 & 1.00\\
8 & 1 & 0.9992 & 0.9992 & 0.0000 & 1.00\\
9 & 1 & 0.8347 & 0.8347 & 0.0000 & 1.00\\
10 & 1 & 0.8314 & 0.8314 & 0.0000 & 1.00\\
11 & 1 & 0.9081 & 0.9081 & 0.0000 & 1.00\\
12 & 1 & 0.8745 & 0.8745 & 0.0000 & 1.00\\
13 & 2 & 0.8598 & 0.4604 & 0.1595 & 1.00\\
14 & 2 & 0.8598 & 0.4604 & 0.1595 & 1.00\\
15 & 1 & 0.9445 & 0.9445 & 0.0000 & 1.00\\
16 & 1 & 0.9903 & 0.9903 & 0.0000 & 1.00\\
17 & 1 & 0.9991 & 0.9991 & 0.0000 & 1.00\\
18 & 1 & 0.9931 & 0.9931 & 0.0000 & 1.00\\
19 & 1 & 0.9012 & 0.9012 & 0.0000 & 1.00\\
20 & 1 & 0.5417 & 0.5417 & 0.0000 & 1.00\\
21 & 1 & 0.9922 & 0.9922 & 0.0000 & 1.00\\
22 & 1 & 0.8743 & 0.8743 & 0.0000 & 0.40\\
23 & 1 & 0.9169 & 0.9169 & 0.0000 & 0.80\\
24 & 2 & 0.9094 & 0.6155 & 0.0864 & 0.40\\
25 & 2 & 0.9214 & 0.9040 & 0.0003 & 0.70\\
\bottomrule\end{tabular}
\end{table}\end{minipage}\par\medskip
\noindent\begin{minipage}{\columnwidth}
\section{Corpus D Query Results}\label{app:D}
\begin{table}[H]\centering\footnotesize\setlength{\tabcolsep}{4pt}
\caption{Corpus D query-level results.}\label{tab:queryD}
\begin{tabular}{rrrrrr}\toprule Q & $k(q)$ & $SS_1$ & $SS_\mu$ & $SS_\sigma$ & ACS\\\midrule
1 & 4 & 0.9980 & 0.9858 & 0.0001 & 0.60\\
2 & 5 & 0.9988 & 0.9884 & 0.0001 & 1.00\\
3 & 5 & 0.8920 & 0.8654 & 0.0008 & 1.00\\
4 & 5 & 0.9920 & 0.7794 & 0.0531 & 1.00\\
5 & 3 & 0.9826 & 0.9735 & 0.0002 & 0.80\\
6 & 5 & 0.9923 & 0.9845 & 0.0002 & 1.00\\
7 & 5 & 0.9541 & 0.7881 & 0.0161 & 0.80\\
8 & 4 & 0.9992 & 0.8578 & 0.0597 & 1.00\\
9 & 5 & 0.8730 & 0.6984 & 0.1047 & 0.41\\
10 & 4 & 0.9488 & 0.8800 & 0.0040 & 1.00\\
11 & 5 & 0.9855 & 0.8902 & 0.0090 & 1.00\\
12 & 6 & 0.9494 & 0.6691 & 0.1173 & 1.00\\
13 & 6 & 0.9919 & 0.7117 & 0.0803 & 1.00\\
14 & 5 & 0.9553 & 0.7558 & 0.1078 & 1.00\\
15 & 4 & 0.9855 & 0.9697 & 0.0003 & 1.00\\
16 & 6 & 0.9966 & 0.9659 & 0.0028 & 1.00\\
17 & 5 & 0.9992 & 0.9966 & 0.0000 & 1.00\\
18 & 3 & 0.9986 & 0.9973 & 0.0000 & 1.00\\
19 & 4 & 0.9497 & 0.9175 & 0.0005 & 1.00\\
20 & 4 & 0.8013 & 0.6150 & 0.0124 & 1.00\\
21 & 4 & 0.9940 & 0.9874 & 0.0001 & 0.90\\
22 & 5 & 0.9960 & 0.8719 & 0.0109 & 1.00\\
23 & 5 & 0.9169 & 0.7626 & 0.0089 & 0.80\\
24 & 5 & 0.9068 & 0.7748 & 0.0528 & 1.00\\
25 & 5 & 0.9556 & 0.9270 & 0.0007 & 0.80\\
\bottomrule\end{tabular}
\end{table}\end{minipage}\par\medskip

\end{document}